# DEBIASED MACHINE LEARNING FOR ESTIMATING THE CAUSAL EFFECT OF URBAN TRAFFIC ON PEDESTRIAN CROSSING BEHAVIOR

**Kimia Kamal**
PhD Student
Laboratory of Innovations in Transportation (LiTrans)
Toronto Metropolitan University
Toronto, Canada
Email: kimia.kamal@ryerson.ca

**Bilal Farooq**
Associate Professor
Laboratory of Innovations in Transportation (LiTrans)
Toronto Metropolitan University
Toronto, Canada
Email: bilal.farooq@ryerson.ca

Word Count: 6438 words + 4 table(s) × 250 = 7438 words

**ABSTRACT**
Before the transition of AVs to urban roads and subsequently unprecedented changes in traffic conditions, evaluation of transportation policies and futuristic road design related to pedestrian crossing behavior is of vital importance. Recent studies analyzed the non-causal impact of various variables on pedestrian waiting time in the presence of AVs. However, we mainly investigate the causal effect of traffic density on pedestrian waiting time. We develop a Double/Debiased Machine Learning (DML) model in which the impact of confounders variable influencing both a policy and an outcome of interest is addressed, resulting in unbiased policy evaluation. Furthermore, we try to analyze the effect of traffic density by developing a copula-based joint model of two main components of pedestrian crossing behavior, pedestrian stress level and waiting time. The copula approach has been widely used in the literature, for addressing self-selection problems, which can be classified as a causality analysis in travel behavior modeling. The results obtained from copula approach and DML are compared based on the effect of traffic density. In DML model structure, the standard error term of density parameter is lower than copula approach and the confidence interval is considerably more reliable. In addition, despite the similar sign of effect, the copula approach estimates the effect of traffic density lower than DML, due to the spurious effect of confounders. In short, the DML model structure can flexibly adjust the impact of confounders by using machine learning algorithms and is more reliable for planning future policies.

*Keywords*: Double machine learning, Copula-based Model, Causal inference, Machine learning, Pedestrian wait time.

**INTRODUCTION**
In the near future, with the introduction of Automated Vehicles (AVs) to urban roads and the presence of different types of vehicles, the interaction between road users is expected to change. Among them, pedestrians are classified as the most vulnerable road users and their behavior analysis plays a crucial role in the implementation of transportation policies. Waiting time is an initial part of pedestrian crossing behavior and represents the duration between arrival at a sidewalk and starting a crossing event. Generally, pedestrians decide to start crossing a street, if they feel comfortable and find the road without any dangers or risk of an accident, meaning safer conditions for crossing. On the one hand, this duration guarantees safe crossing and on the other hand, waiting time may be considered as a delay and exerts massive pressure on pedestrians. Therefore, analyzing the effect of transportation policies on pedestrians' wait time is of vital importance in the presence of AVs that pedestrians will face unprecedented traffic conditions. In recent years, researchers have demonstrated that various factors, particularly road characteristics and traffic conditions influence waiting time on the sidewalk in the presence of AVs. Kalatian and Farooq (*1*) and Kamal and Farooq (*2*) showed that traffic density is a determining factor in pedestrian crossing behavior in such a way that higher traffic density highly significantly affects longer waiting time. Indeed, as pedestrians constantly try to find an appropriate gap for crossing, higher density decreases the opportunity for pedestrians to cross safely, except in a fully traffic congestion condition. Hence, the result of these recent studies highlighted that traffic control systems, leading to lower traffic density, can be contemplated as one policy for controlling unexpected delays coming from traffic conditions in the presence of AVs.

Nevertheless, in addition to traffic density, some personal characteristics such as a pedestrian stress level may influence their waiting time at a sidewalk of mid-block crosswalk (*3*). In other words, this factor might confound the effect of traffic control policies on pedestrian wait time, leading to biased estimation. For instance, for implementing a control system of traffic density, based on conducted studies we may mistakenly conclude this policy can decrease waiting time considerably, while the main reason that some people wait shorter on the sidewalk can be rooted in their characteristics. In other words, they usually feel less stress and can immediately find an appropriate gap and cross the street. Therefore, to evaluate the policies related to pedestrians waiting time in the presence of AVs, we must account for the impact of some variables in the analysis that simultaneously have a direct effect on the policy and the outcome. These variables are known as confounding variables or confounders.

Different methods have been employed by researchers to model multiple dependent variables and address confounders in various fields of study. In transportation, in recent years, the impact of contributing factors and policies have been assessed by using more accurate probabilistic behavioral models such as copula-based joint models, promising machine learning algorithms and deep neural networks (*1, 2, 4–6*). However, these models are criticized for learning spurious relationships, resulting in biased estimates and invalid causal conclusions. This issue is primarily attributed to the lack of causal inference in travel behavioral models (*7*), as many questions in the policy assessment process are causal in nature. In contrast, causal inference techniques allow us to purely evaluate the effect of policies and overcome the effect of confounding variables such as pedestrian stress level on waiting time. Broadly, travel behavioral models analyze associations between a dependent variable and various explanatory variables and are not mainly applied for determining the causal impact of a policy on an outcome, as these models are not able to address the spurious effect of confounders.

The objective of this study is to investigate the impact of control of traffic density on pedestrian wait time on the sidewalk of mid-block crosswalk by applying two different model structures. We first develop a copula-based joint model of stress level and waiting time to analyze the impact of density on pedestrian wait time. In fact, we try to jointly model two main components of crossing decision process with the intention of evaluating the effect of the traffic density by an accurate model representing the real world. Secondly, we employ Double/Debiased Machine Learning (DML), which to the best knowledge of authors, is a novel approach for determining the causal effect between a policy and an outcome. More precisely, the DML framework is a machine learning-based causal model in which the spurious impact of confounders is addressed (*8*). As in recent years, machine learning algorithms provide the feasibility of capturing real-world behavior, we try to infer causality between traffic density and pedestrian waiting time by using the DML model structure, a machine learning-based causal model.

The rest of this paper is organized as follows. First, a brief review of the literature on pedestrian waiting time studies, causality analysis in transportation and machine learning-based causal models are provided. It is followed by the methodology section in which the structure of copula approach and DML model is described in detail. Then, a description of the data and the results of models are provided. Finally, in the conclusion section, some suggestions and future directions of the study are proposed.

## BACKGROUND
In this section, we first present a general overview of the research on pedestrian wait time. Then a brief review of causality analysis in travel behavior modelling in transportation and machine learning based-causal model is provided in the following sections.

### Interaction between pedestrians and automated vehicles
Pedestrian crossing behavior analysis in the presence of AVs has received relatively less attention in the literature. Gender, age, home location and people's awareness towards innovation were observed as contributing factors in the acceptance of AVs (*9, 10*). Furthermore, pedestrians' trust in vehicle technology and road design is a determining factor in pedestrian crossing studies (*11*). In very recent years, Kalatian and Farooq tried to bridge the gap between pedestrian crossing behavior and mixed traffic condition (*1*). In contrast to previous studies, they mainly focused on road attributes and traffic conditions such as street width, road speed, traffic flow, etc., by applying a novel virtual reality tool for data collection. In their study, they employed a deep neural network version of CPH for analyzing the impact of various variables and showed that lacking walking habit, older pedestrians, wider lane widths, higher traffic density and poor sight distance are lead to longer waiting time (*1*). Previously, for the first time, we investigated the ordinal categories of pedestrian wait time in the presence of AVs by proposing a fully interpretable deep learning-based ordinal regression model (*2*). Similarly, our results highlighted the significant impact of road-related variables and traffic conditions. We particularly emphasized the crucial role of traffic density on pedestrian crossing behavior prediction. In another study, we focused on pedestrian stress levels when they are waiting on a sidewalk in the presence of AVs (*12*). In this study, we also used a virtual reality dataset for analyzing the effect of road attributes, traffic conditions and environmental situations on the ordinal form of pedestrian stress level. Our previous studies concluded that longer waiting time is probably rooted in the higher stress level of pedestrians.

**Causality analysis in travel behavior modelling**
Over the past several decades, travel behavior models have been used in various transportation studies with the intention of evaluating the impact of various explanatory variables related to transportation policies and plans. As the policy assessment process are causal in practice, some concepts from causal inference academic disciplines have been applied in transportation studies.

In transportation, the propensity score method is the most traditional approach for estimating the effect of a policy by analyzing the variables which define the characteristics of policies (*13, 14*). In this approach, the impact of a policy is easily measured by comparing the value of outcomes between individuals exposed to the policy, known as a treatment group, and those who are considered as a control group. As in observational studies, the assignment of policies is generally not random, the propensity score method allows us to mimic a quasi-random experiment, by considering two types of groups. In fact, after pairing or grouping observations with similar propensity score, any difference between two groups with respect to outcome can accurately be traced back to true treatment effect, not a difference arising from self-selection phenomenon or in other words confounding variables. In transportation, the phenomenon in which individuals' tendency toward a certain kind of behavior determines their decision is known as a self-selection problem. Sample selection is another popular approach used for addressing the self-selection problem. The standard sample selection approach consists of two equations 1) a selection equation 2) an outcome equation which is specified based on travel behavior indicators of interest like household vehicle miles of travel (VMT), travel mode choice, telecommuting frequency, etc (*15*). A considerable difference between propensity score and sample selection methods is that the latter approach also captures the correlation between unobserved variables of the selection equation and outcome equations. Copula approach was also widely used in the literature for these problems. The copula-based joint model is one type of sample-selection method in which the restrictive assumption of standard sample selection approach, which is bivariate normally distributed errors between the selection equation and outcome equations, is relaxed (*16*). In other words, the copula approach allows us to model a stochastic correlation between multiple dependent variables without imposing distribution assumptions on error terms. In recent years, this approach is widely used in the transportation literature to jointly model different random variables such as mode choice and travel distance (*4*), vehicle type and miles of travel (*17*), collision type and crash injury severity (*18*).

Although in this study, our focus is on addressing observed confounders, it is worth mentioning that eliminating the impact of unobserved confounding variables recently have been received attention among researchers in transportation. Among them, this concept is popular as an endogeneity problem. In econometrics, endogeneity refers to an issue in which an explanatory variable is correlated with the error term. However, in causal inference, this issue is described as a situation in which some variables afflict the causal relationship between a treatment and an outcome. In travel behavior modeling, particularly instrumental variable methods are used to address this issue (*19*).

**Machine learning-based causal models**
In recent years, several fields of science and engineering have demonstrated significant improvements in the estimation and evaluation by promising performances of machine learning algorithms. Despite considerable success in the development of advanced travel behavior models by applying machine learning algorithms and correctly specifying behavioral assumptions (*2, 5, 6*), these mod-els are just able to consider associations between a dependent variable and explanatory variables. However, associations between a dependent variable and explanatory variables are not exactly equal to true or causal effect (*7*). In fact, most machine learning algorithms aim to predict outcomes rather than understanding causality, while prediction is not merely enough for describing decision-making process. However, in the causal inference discipline, the main goal is to assess the causal effect of a potential cause, known as a policy in transportation, on an outcome. Therefore, in very recent years, there have been massive efforts to propose machine learning-based causal models so as to enjoy the advantages of both causal inference discipline and machine learning algorithms. In fact, not only does introduction of causality into machine learning algorithms provide the feasibility of capturing the relationship between input and output directly from the data, but also this model allows us to eliminate the effect of a confounder, resulting in a true cause-effect relationship. Another important property of machine learning-based causal models is that these models do not suffer from lack of interpretability in such a way that modelers are able to analyze the effect of causes or policies.

Lee et al. (*20*) tried to improve the performance of propensity score method by using machine learning models. The Causal Bayesian Network is a sophisticated approach which estimates causal relationships between variables of data based on a directed causal graph where nodes represent random variables and edges represent the relationship between them. This method is able to simultaneously discover multiple true cause-effect (*21*). Causal forest method is an extended version of random forests model structure to measure the causal effect of an intervention on an outcome. The main difference between causal forest and random forest is the splitting criterion (*22*). Based on the potential outcome concept, the prediction of each policy effect is equal to the difference of an average outcome between the treatment and control groups (*23*). Therefore, in causal forest splitting criterion aims to maximize the difference across splits in the determining of each policy effect. Double/Debiased Machine Learning (DML) is another novel approach proposed by Chernozhukov et al. (*8*). The DML model is able to capture the relationship between a policy of interest and confounding variables. Therefore, this model structure accounts for confounders and is able to identify true causal effect of a policy on an outcome. Interestingly, DML enjoys machine learning algorithms for estimating the dependency between both the outcome and confounder and the policy and confounding variables. In fact, modelers are interested in applying machine learning algorithms to infer causality due to some reasons. First, machine learning models generally perform better than traditional statistical methods in recent years, that modelers usually use big datasets. Secondly, machine learning frameworks relax strict assumptions of traditional models and learn the relationship between variables directly from data. Indeed, DML is structured based on machine learning algorithms and is classified as an unbiased estimator. It is of note that in addition to causal linear relationship, DML is able to capture fully non-linear policy effects. For instance, the Double Random Forest is a combination of causal forests and double machine learning (*24*). In the next section, the structure of DML is completely described.

## METHODOLOGY

In this study, we aim to analyze the effect of traffic density on pedestrian wait time. First, we develop a copula-based joint model of stress level and waiting time for addressing the effect of selection bias. Copula functions allow us to associate the function of main components of pedestrian crossing behavior. Then, a machine learning-based causal model named Double/Debiased Machine Learning (DML) model is employed for accounting confounding variables. In our case study, the pedestrians stress levels can be one confounder. More precisely, although some pedestrians may make a decision on the sidewalk based on their stress level, we may misleadingly consider traffic density as the main reason of pedestrian wait time. Eventually, we compare the impact of traffic density obtained from both model structures.

**Copula approach**

Copula is a function that associates the functions of several random variables by their marginal distributions and defines them as a multivariate dependent distribution. In our analysis, there are two dependent variables, pedestrians stress level and pedestrian waiting time.

First, we model pedestrian stress level by the binary logit model as follows:

$$U_q = \beta X_q + \varepsilon_q \qquad (1)$$

Where q denotes the pedestrians and $U_q$ is the propensity for a low or high level of stress in pedestrians' body. $X_q$ is the vector of explanatory variables, $\beta$ is the vector of parameters to be estimated, and $\varepsilon_q$, is the error term following Gumbel distribution. If pedestrians perceive dangers of interaction with vehicles or the risk of an accident, increasing the level of stress would be the reaction of their body. Hence, in the case of pedestrian behavior analysis, pedestrians feel a high level of stress if the value of the utility function, which represents the propensity of their body towards stress, is positive. Therefore, experiencing a high or low level of stress for each pedestrian can be defined by equation (2):

$$r_q = \begin{cases} 1, & \text{if } U_q > 0 \\ 0, & \text{if } U_q \leq 0 \end{cases} \qquad (2)$$

For each pedestrian, if the stress level is high, $r_q = 1$, otherwise $r_q = 0$. Moreover, pedestrian waiting time is modeled by the ordered logit model. In the real world, people usually describe their waiting time or delay on a sidewalk by low, medium or high phrases. Therefore, we use the ordered logit model for evaluating pedestrian waiting time in the copula-based joint model as follows:

$$V_q^* = \gamma Z_q + \eta_q \qquad (3)$$

$$V_q = k \text{ if } \delta_{k-1} < V_q^* \leq \delta_k$$

Where $V_q$ represents the ordinal response and $V_q^*$ is the latent variable which is assumed to be a linear function of explanatory variables ($Z_q$) with associated parameters ($\gamma$), which remain constant between ordinal categories. In equation (3), $\eta_n$ is the random error component of the utility function such that the logistic distribution is the most common choice for $\eta_n$. A set of thresholds ($\delta_k$) associates the latent variable ($V_q^*$) to ordinal response ($V_q$). It is of note that thresholds must satisfy the constraint $\delta_1 \leq \delta_2 \leq ... \leq \delta_k$ in which k is the index of pedestrian waiting time ($k = 1$ for low waiting time, $k = 2$ for medium waiting time and $k = 3$ for high waiting time). Typically, it is assumed that $\delta_0 = -\infty$ and $\delta_3 = +\infty$.

In our case study, the probability that each individual feels a high level of stress and chooses low, medium or high waiting time can be obtained through equation (4).

$$\begin{aligned}
Pr(r_q = 1, V_q = k) &= Pr(U_q > 0, \delta_{k-1} < V_q^* < \delta_k) \\
&= Pr(\varepsilon_q > -\beta X_q, \eta_q < \delta_k - \gamma Z_q) - pr(\varepsilon_q > -\beta X_q, \eta_q < \delta_{k-1} - \gamma Z_q) \\
&= Pr(\eta_q < \delta_k - \gamma Z_q) - Pr(\varepsilon_q < -\beta X_q, \eta_q < \delta_k - \gamma Z_q) \\
&\quad - Pr(\eta_q < \delta_{k-1} - \gamma Z_q) + Pr(\varepsilon_q < -\beta X_q, \eta_q < \delta_{k-1} - \gamma Z_q) \\
&= G(\delta_k - \gamma Z_q) - Pr(\varepsilon_q < -\beta X_q, \eta_q < \delta_k - \gamma Z_q) \\
&\quad - G(\delta_{k-1} - \gamma Z_q) + Pr(\varepsilon_q < -\beta X_q, \eta_q < \delta_{k-1} - \gamma Z_q)
\end{aligned} \quad (4)$$

This probability function relies on the joint distribution of the two random variables; therefore, by using the copula function, equation (4) can be written as follows:

$$\begin{aligned}
Pr(r_q = 1, V_q = k) &= G(\delta_k - \gamma Z_q) - C_\theta[F(-\beta X_q), G(\delta_k - \gamma Z_q)] \\
&\quad - G(\delta_{k-1} - \gamma Z_q) + C_\theta[F(-\beta X_q), G(\delta_{k-1} - \gamma Z_q)]
\end{aligned} \quad (5)$$

Where $G(.)$ and $F(.)$ are marginal distributions of random error terms of $\eta_q$, $\varepsilon_q$ respectively and $\theta$ is copula parameters capturing the correlation between random variables of models. Similarly, the probability that each individual feels a low level of stress and chooses different categories of waiting time is computed by equation (6).

$$Pr(r_q = 0, V_q = k) = C_\theta[F(-\beta X_q), G(\delta_k - \gamma Z_q)] - C_\theta[F(-\beta X_q), G(\delta_{k-1} - \gamma Z_q)] \quad (6)$$

For the estimation of parameters of models, the Maximum likelihood method is employed. There are different copula functions in the literature by which we are able to consider different types of dependency between random variables. Detailed information about description and comparison of different copula functions is provided Bhat and Eluru (*16*). Different copula functions can be compared based on the Bayesian Information Criterion (BIC) and is equal to equation (7) as follows.

$$BIC = -2\ln(L) + k\ln(n) \quad (7)$$

Where $\ln(L)$ is the maximum log-likelihood function, $k$ is the number of estimated parameters, and $n$ is the total number of observations.

**Double/Debiased machine learning model**
Double/Debiased Machine Learning (DML) is a causal machine learning model that combines two main predictive models 1) outcome equation predicting the relationship between the outcome and variables and 2) policy equation which models the effect of confounding variables on the policy (*25*). The DML model is based on the doubly robust method (*26*) and the main assumption of this model structure is unconfoundedness assumption in such a way that all potential confounders are observed. DML framework allows us to account for high-dimensional confounders and also model the effect of variables on the treatment and outcome by supervised machine learning methods. Moreover, this model structure provides the feasibility of evaluating binary or continuous form of policies (*27, 28*). The structural equation of partially-linear DML is defined by:

$$Y = D\alpha + g(W) + \zeta, \ \mathbf{E}[\zeta|W, D] = 0 \quad (8)$$

$$D = f(W) + v, \quad \mathbf{E}[v|W] = 0 \tag{9}$$

Where $Y$ is pedestrian wait time, $D$ is a binary form of traffic density, and $W$ is a vector of variables influencing both the policy and outcome. For instance, in our case study, we believe that not only does pedestrian stress level affect their waiting time but also has an effect on perceiving the traffic density. In fact, pedestrians with a different state of stress may find an appropriate gap for crossing at a different time due to their perception of traffic density, if we assume that all other factors are constant. Moreover, $g$ and $f$ are called nuisance functions that estimate the dependency between both the outcome and confounders and the policy and confounding variables respectively.

The equation (9) helps us to overcome the bias rooted in confounders in such a way that we consider the effect of these variables on both policy and outcome models. Interestingly, there is no strict assumptions on the model specification of $g$ and $f$ and we are able to estimate the non-parametric effect of confounders using machine learning methods.

For estimation, First, we separately predict $D$ and $Y$ based on confounders using machine learning algorithms. Then, we can compute the residuals of two models as follows:

$$\tilde{Y} = Y - g(W) \tag{10}$$

$$\tilde{D} = D - f(W) \tag{11}$$

Finally by regressing the residuals of two models, $\alpha$ or average policy effect is estimated:

$$\tilde{Y} = \tilde{D}.\alpha + \zeta \tag{12}$$

This method guarantees Neyman orthogonality condition, meaning that the estimate of $\alpha$ is robust to small errors in nuisance functions (*25*). In general, DML model structure is classified as an unbiased and root n-consistent estimator of the average policy effect in the literature, which enjoys advantages in terms of convergence and data efficiency (*29, 30*). In this model structure, the Generalized Method of Moment (GMM) is employed. GMM is a method for estimating parameters, analogous to maximum likelihood but GMM is usually used in semi-parametric models. While the maximum likelihood estimator uses assumptions about the entire distribution, GMM assumes that the full shape of the data distribution may be unknown and uses assumptions about specific moments of the random variables. This assumption are referred to as moment conditions. In fact, moment conditions can be expected values of random variables and GMM method specify the model parameters in such a way that moment condition are closest to true moments. More precisely, Based on the idea of GMM method, the cost function measures the deviation of all moment conditions from true moment. It is of note that Ordinary Least Square (method) is based on GMM method where the error term is normally distributed. Therefore, for estimating the parameters of DML method, the cost function ($J(\alpha)$) is defined as a product of error terms of predicted models as follows:

$$J(\alpha) = (D - f(W)) \times (Y - g(W) - (D - f(W))\alpha) \tag{13}$$

We make this cost function equal to zero. To address the overfitting problem in the sense that no observation is used to for estimating its own nuisance functions, K-fold cross-fitting method is used. We randomly partition our data into K folds and estimate machine learning models for nuisance functions on K-1 folds. Then, the policy effect ($\alpha$) is estimated on fold k left out by using two nuisance models.

**DATA**
In this study, a virtual reality dataset which is collected based on the Virtual Immersive Reality Experiment (VIRE) is used. VIRE is a virtual reality simulation framework providing a perception of being physically present in futuristic scenarios by immersing the participants in an artificial 3D environment (*31*).

In this dataset, pedestrians stress levels and their waiting time are collected in different scenarios in which pedestrians interact with the current traffic condition, fully automated vehicles and the mixed traffic condition. These scenarios were defined based on different controlled factors such as rules and regulations, street characteristics, automated vehicle features, traffic demand and environmental conditions (*3*). Moreover, during experiments, Galvanic Skin Response (GSR) sensors technology was used to measure the relative stress levels of a participant. The GSR sensor uses a small electrical charge to measure the amount of sweat an individual has on their finger (*3*). The greater the charge, the greater the sweat. Detailed information about the definition of scenarios can be found in their study. As the GSR sensor only measures the relative change in the stress of a pedestrian compared to their initial stress, the data used in (*3*) cannot be used directly to compare the stress among the pedestrians or develop models. Therefore, in this study, the mean stress level of the pedestrian in a scenario was normalized according to their minimum and maximum stress level observed in each scenario.

In total, this dataset consists of 1,406 responses for pedestrian crossing behavior analysis. A list of explanatory variables which are used in this study is presented in Table 1. The Jenks Natural Breaks classification method is applied to categorize normalized stress and waiting time into discrete categories (*32*). The chosen method classifies a given data into several groups in such a way as to minimize the variance of members of each group while maximizing the variance between different groups. Regarding normalized stress variable, the data is categorized into two categories, low and high groups, in such a way that pedestrians who relatively feel minor stress level, less than 0.5 are allocated into low categories. Furthermore, the waiting time data is classified into three discrete groups based on the Jenks classification method and logical facts: low: pedestrians who wait less than 5 seconds, medium: a group of pedestrians who has waiting time between 5 to 20 seconds on the sidewalk, and high: pedestrians waiting more than 20 seconds. The frequencies of crossing with different stress level and waiting time are shown in Figure 1.

Concerning traffic density, based on the controlled variables of experiments, density of less than 18.75 is presumed as a policy evaluated in this study. 18.75 (veh/hr/ln) is an average density related to a traffic condition in which road speed and traffic flow are designed 40 (km/hr) and 750 (veh/hr), respectively.

**RESULT**
In this section, first, the results obtained from the copula approach are described. Then, the causal effect between pedestrian wait time and traffic density investigated by the DML model is presented.

TABLE 1: Description of Explanatory Variables

| Variable | Description | Mean | Standard deviation |
|---|---|---|---|
| **_Street attributes_** | | | |
| Low lane width | 1: If the lane width is less than 2.75 meter, 0:otherwise | 0.334 | 0.472 |
| High lane width | 1: If the lane width is greater than 2.75 meter, 0:otherwise | 0.359 | 0.480 |
| Two way with a median | 1: If the road type is two way with median, 0:otherwise | 0.338 | 0.473 |
| Two way | 1: If the road type is two way, 0:otherwise | 0.317 | 0.466 |
| One way | 1: If the road type is one way, 0:otherwise | 0.344 | 0.475 |
| **_Traffic condition_** | | | |
| Mixed traffic condition | 1: If traffic in scenario consists of automated vehicles and human-driven vehicles, 0:otherwise | 0.045 | 0.207 |
| Fully automated condition | 1: If traffic in scenario consists of only automated vehicles, 0:otherwise | 0.925 | 0.263 |
| Human-driven condition | 1: If traffic in scenario consists of only human-driven vehicles, 0:otherwise | 0.029 | 0.170 |
| **_Socio-demographic_** | | | |
| Age 18-30 | 1: If participant' age is between 18 and 30, 0:otherwise | 0.542 | 0.498 |
| Age 30-39 | 1: If participant' age is between 30 and 39, 0:otherwise | 0.339 | 0.474 |
| Age 40-49 | 1: If participant' age is between 40 and 49, 0:otherwise | 0.034 | 0.182 |
| Age over 50 | 1: If participant' age is more than 50, 0:otherwise | 0.084 | 0.278 |
| Female | 1: If participant is female, 0:otherwise | 0.382 | 0.486 |
| Driving license | 1: If participant has a driving license, 0:otherwise | 0.895 | 0.307 |
| No car | 1: If participant has no car in the household, 0:otherwise | 0.302 | 0.345 |
| One car | 1: If participant has one car in the household, 0:otherwise | 0.345 | 0.476 |
| Over one car | 1: If participant has more than one car in the household, 0:otherwise | 0.353 | 0.478 |
| Active mode | 1: If participant use active modes regularly, 0:otherwise | 0.258 | 0.438 |
| Private car mode | 1:If participant use private car regularly, 0:otherwise | 0.332 | 0.471 |
| Public mode | 1: If participant use transit regularly, 0:otherwise | 0.409 | 0.492 |
| VR experience | 1: If participant has VR experience, 0:otherwise | 0.425 | 0.495 |
| **_Environment condition_** | | | |
| Night | 1: If the time of scenario is night, 0:otherwise | 0.371 | 0.483 |
| Snowy | 1: If the weather of scenario is snowy, 0:otherwise | 0.293 | 0.455 |

**Copula Approach**
*Model performance*
In this study, we evaluate the performance of three flexible copula functions, Frank, FGM and Gaussian, which are able to capture both positive and negative dependency between pedestrian stress level and waiting time. The performances of these copula functions are assessed based on their properties and BIC which is a common measure for comparing different copulas. The results showed that Frank function outperforms other copula functions with lower BIC, approximately 2738.56. Furthermore, we compare the performance of Frank function with product copula as an independent model structure. Interestingly, Frank copula function obtains a better statistical fit with lower BIC compared to the product function (with BIC of 2954.57). Moreover, the statistical significance of the copula parameter highlights that there is a negative dependency between pedestrian stress level and their waiting time.

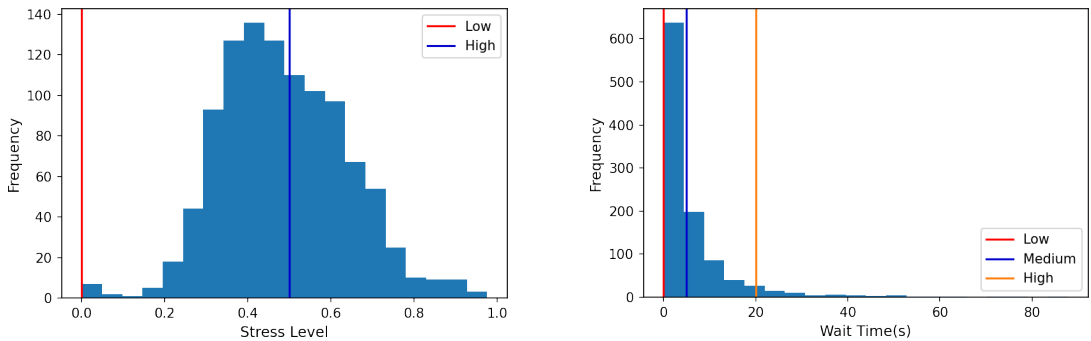

**FIGURE 1**: Wait time and stress level frequency

The significant negative sign of copula parameter reveals that we should consider the effect of stress levels on pedestrian waiting time in the behavioral analysis and policy evaluations. However, the results of Frank copula show that if pedestrians feel a higher level of stress probably due to their personal characteristics, road attributes or traffic condition, they are likely to wait shorter on the sidewalk. This result is not in line with real-world behavior. In fact, individuals who are under more stress usually hesitate to cross the street immediately. This result may be traced back to the lack of causal inference concept in travel behavior models (*7*). In other words, copula model structure is only capable of considering the association between variables.

*The Impact of Explanatory Variables*
The results obtained from the Frank copula function are presented in Table 2. The positive coefficient of road types shows pedestrians likely find the one-way roads safer for crossing in the presence of AVs but their waiting time are getting higher. Moreover, in this study, we surprisingly observe the negative effect of wider lane width on pedestrian stress levels. While, the effect of this variable on pedestrian waiting time is reasonably similar to our previous study. Results reveal that traffic density affects only one component of pedestrians behavior crossing, waiting time. Logically, the result shows that lower density leads to lower waiting time. In fact, in this traffic condition, people usually find an appropriate gap sooner for crossing and they tend to wait shorter in the lower traffic density.

Regarding traffic conditions, with the introduction of AVs into urban roads, pedestrians probably will feel less stress levels compared to the current traffic condition and in the mixed automated situation pedestrians' preference towards higher waiting times will be increased. Interestingly, in terms of stress level, similar results were obtained in our previous study. In fact, this result highlights the importance of general trust in AVs and braking systems of vehicles. In other words, this result demonstrates that more trust in AVs leads to lower stress level and also underscores the effect of considering confounder variables in the process of transportation polices

Kamal and Farooq       13

**TABLE 2**: Results of Copula Joint Model

| Variable | Stress Level | | Wait Time | |
|---|---|---|---|---|
| | Value | t-stats | Value | t-stats |
| ***Street characteristics*** | | | | |
| High lane width | -0.647 | -3.739 | 0.383 | 2.513 |
| Two way with a median | 0.226 | 1.408 | -0.726 | -4.817 |
| Density | - | - | -0.630 | -1.542 |
| ***Traffic condition*** | | | | |
| Mixed traffic condition | -1.584 | -3.404 | 0.568 | 1.047 |
| Fully automated condition | -1.821 | -7.072 | 1.131 | 2.434 |
| ***Environmental variables*** | | | | |
| Snowy | 0.365 | 2.258 | - | - |
| ***Socio-demographic*** | | | | |
| Age 30-39 | - | - | -1.389 | -8.549 |
| Age 40-49 | 0.977 | 2.460 | - | - |
| Age over 50 | 1.732 | 6.896 | - | - |
| Female | 0.738 | 4.148 | 1.194 | 7.852 |
| Driving license | 0.361 | 1.549 | - | - |
| Over one car | -0.378 | -2.100 | - | - |
| One car | - | - | 0.648 | 4.374 |
| Active mode | -0.259 | -1.446 | -0.598 | -3.176 |
| Public mode | 0.237 | 1.414 | -1.170 | -6.888 |
| Threshold1 | 0.316 (1.824) | | | |
| Threshold2 | 2.964 (15.578) | | | |
| Copula parameter | -0.738 (-2.101) | | | |
| No. observation | 1046 | | | |
| No. parameters | 26 | | | |
| Log-likelihood | -1278.896 | | | |
| BIC | 2738.563 | | | |

evaluation before the introduction of AVs into urban roads. More precisely, people's attitude may confound the effect of policies on the outcome of interest, and we must account for confounders in modeling.

Another noteworthy aspect of this study which is rooted in virtual reality dataset is evaluation of the effect of environmental condition, snowy and night conditions. Snowy weather increases the probability of a higher stress level, since harsh weather condition afflicts pedestrians' vision and their ability for detecting distance, vehicle speed, etc.

The Frank copula model significantly estimated the effect of some individuals' characteristics. The most noticeable results of our analysis is about people's lifestyle. Our result indicates that in the face of AVs, pedestrians who prefer to regularly use public transportation will probably find the future condition more stressful, but they wait shorter on the sidewalk. In our previous studies, we traced back these results to the regular schedule of public transportation, causing people to behave hurriedly while they are under more pressure in comparison to people who usually use a private car. We also observe that the probability of having higher stress considerably decreases if pedestrians usually choose active modes and they prefer to wait shorter on a sidewalk. Surprisingly, in terms of waiting time, we found a completely opposite result in our previous work by applying a the deep learning-based ordinal logit model.

Although the copula approach gives us this opportunity to model two dimensions of pedestrian crossing behaviors jointly, this model structure lacks considering the effect of confounders in estimation and addressing spurious relationships. Copula joint model only allows us to consider a joint decision process to reduce selection bias and analyze associations between variables that can be classified under causal association or non-causal association. However, for policy evaluation, modelers require causal-based models. Based on the interpretation of the Frank copula model results, we can conclude that some variables like pedestrians' characteristics may play as confounding variables in the decision of crossing a street. Therefore, in the next section, we strive to overcome the effect of confounders by using a causal machine learning model.

**Double/Debiased Machine Learning**
As in previous studies, we found that traffic density significantly affects pedestrian wait time (*1, 2*), in this study, we mainly focus on the causal effect of a binary form of traffic density on pedestrians waiting time so as to prevent an unexpected increase in pedestrians' delay, which may lead to unbearable situation and endanger pedestrians.

However, predominant travel behavior analysis such as copula approach ignore the effect of confounding variables and modelers barely analyze the associational relationship between a dependent variable and various explanatory variables. While, the effect of individuals' preference and their characteristics is an obvious example of a confounding variable that not only may confound the effect of other variables on the dependent variable of interest but also can directly influence the dependent variable. To determine the causal effect of traffic density on pedestrian wait time and eliminate spurious impacts of confounding variables, the Double/Debias machine learning (DML) model is employed. The principle assumption of the DML model is that all confounding variables are observed (*25*). Therefore, we first try to specify a set of variables confounding the effect of traffic density on pedestrian wait time. We define a set of confounding variables based on our prior knowledge and previous conducted studies (*1, 2*). Weather conditions and the presence of AVs may influence the average number of vehicles that occupy one kilometer of a road lane. Furthermore, it is self-evident that individual' characteristics and their state of stress may influence their perception of density and as a result, confounds the effect of our policy of interest. To clarify this issue, in the real world, if ordinary people are asked about the level of service of streets which depends on traffic density, it is highly likely that their response would be different. In fact, they assess the condition of traffic based on their personal characteristics. For instance, if they are female or usually feel great stress in the face of vehicles, their response and subsequently their behavior reflect their characteristics. Nevertheless, the results obtained from our previous studies and copula-based joint model show that there is a strong relationship between aforementioned variables and pedestrian waiting time. Hence, in this study, individual' characteristics, snowy weather and traffic condition are considered as confounders since not only they may confound the effect of traffic density (policy) on pedestrian wait time (outcome) but also can directly influence pedestrian wait time.

**TABLE 3**: Hyperparamater of Random Forest Estimators

| Parameters | Pedestrian wait time model | traffic density model |
|---|---|---|
| Number of trees | 100 | 200 |
| Minimum number of samples at each node | 10 | 10 |
| Minimum number of samples at a leaf node | 1 | 3 |
| Number of features | Sqrt | Sqrt |
| Maximum depth of the tree | None | None |
| Bootstrap | True | True |

**TABLE 4**: Causal Effect of traffic density

| Paramater | Value | Standard error | Confidence interval | Cost value |
|---|---|---|---|---|
| Density | -1.115 | 0.132 | [-1.123, -1.107] | 9.099 |

In this study, we develop a partially-linear DML in which the effect of traffic density (policy) on pedestrian wait time (outcome) is considered linear. However, Random Forest estimators are used for modeling the effect of other variables influencing both pedestrians wait time and traffic density. In this model structure, machine learning algorithms allow us to model the effect of confounders by non-parametric and more flexible functions. The performance of machine learning algorithms noticeably relies on the setting of hyperparameters. Based on 10-fold cross-validation, the optimum values for hyperparameters of random forest estimators are presented in Table 3. It is of note that the DML model structure gives us this opportunity to consider the effect of traffic density as a function of road attributes. In other words, instead of a constant effect, different model classes can be assumed for $\alpha$, such as linear or non-parametric functions. In this study, we analyzed different types of DML. However, we observed that the partially-linear DML model with a constant effect of theta outperforms other models. The results obtained from the partially-linear DML model as a final causal machine learning model are presented in Table 4. This Table shows the causal effect of traffic density on pedestrian wait time, standard error and the related confidence interval of our estimate.

To ensure that the bias rooted in confounders asymptotically disappears, the standard error and, thus, the confidence interval should be as small as possible. It is worth mentioning that in contrast to copula method in which we analyze the effect of distinct explanatory variables on pedestrian wait time, in this causal model structure, our focus is only on the impact of traffic density on pedestrian wait time not other variables. Therefore, in this study, the results of the copula approach and DML are compared based on the effect of traffic density. Based on the information of Table 4, traffic density has a negative causal effect with a value of -1.115, meaning that pedestrians tend to wait shorter in the lower traffic density. Although the sign of the average effect is similar to the result of copula approach, but there is a considerable difference between the value obtained from these two methods. Despite jointly modeling two main components of crossing behavior, the copula approach is not able to eliminate the effect of confounders; accordingly, the effect of density on pedestrian waiting time is different between the two approaches. In the copula model, the result cannot be completely traced back to the policy and it may be rooted in pedestrians' preferences and characteristics. While DML exactly estimates the causal effect of density by adjusting for confounders flexibly. Therefore, this model structure can be more reliable for planning future policies in the presence of AVs. In other words, if we assess the effect of traffic density on pedestrian crossing waiting time based on the biased results of the copula joint model, we undermine the effect of this policy. Nonetheless, DML demonstrates that controlling traffic density has averagely higher impact on decreasing pedestrian wait time when all other variables are kept constant. In addition, the standard error term of density parameter is lower than copula approach and the confidence interval is considerably small in such a way that, in 95% level of confidence, the mean parameter is between -1.123 and -1.107.

**CONCLUSION**
In this study, the impact of traffic density on pedestrian wait time in the presence of Automated Vehicles (AVs) is investigated. Although previous studies in this area are conducted by non-causal survival analysis or discrete choice models, in this study, we strive to determine the causal impact of a binary form of traffic density on pedestrian wait time by accounting for confounding variables. Confounders are some variables in behavioral analysis that simultaneously have a direct effect on the policy and the outcome.

In this study, we applied two different model structures for the policy evaluation, the Copula-based joint model and Double/Debiased Machine Learning (DML) model. The copula approach is widely used in transportation literature for addressing self-selection problems. Therefore, we develop a copula-based joint model of pedestrian stress level and waiting time so as to provide a more representative model of real-world behavior. However, this model structure is not capable of adjusting for confounders and estimating causal effect of traffic density on pedestrian wait time. In fact, like other travel behavior models, the copula approach also only considers associations between variables. Hence, to determine causal impact of traffic density on pedestrian wait time, the Double/Debiased Machine Learning (DML) model is employed. Interestingly, The DML model is able to address spurious or non-causal relationships rooted in dependency between a policy of interest and confounding variables. Interestingly, DML is structured based on machine learning algorithms for accounting for confounders.

A virtual reality dataset with 1,406 responses is used for assessing the effect of traffic density. The results of the copula approach and DML are compared based on the effect of lower traffic density on pedestrian wait time. The results show that this variable has a negative effect on pedestrian waiting time in both models, meaning that the lower density leads to lower waiting time. However, copula approach undermines the effect of this policy on the outcome of interest in this study. The standard error term of density variable in DML model is lower than copula approach and the confidence interval is considerably small. In fact, Despite jointly modeling two main components of crossing behavior, the copula approach suffers from lack of ability of adjusting for the effect of confounders. In addition, unbiased DML estimator enjoys advantages of machine learning algorithms for modeling the impact of confounders.

There are some practical implications and useful policy recommendations that can be derived from the results of this study, which can be of use to urban planners and policy makers. Based on our result, this is of importance considering that controlling traffic density cause pedestrians to find an appropriate gap sooner for crossing, resulting in shorter waiting times in the presence of automated vehicles. This observation requires the attention of city planners and decision makers to make modifications needed in urban roads and Travel Demand Management (TDM) techniques. Generally, in current traffic condition, transportation planners opine that using current capacity of the network is more efficient rather than constructing road infrastructure to reduce traffic condition. TDM is the application of policies by which travel demand is reduced or redistributed to the part of the network which does not suffer from traffic congestion. According to the result of our studies, these techniques will be used for reducing traffic density of roads where high traffic density may increase pedestrian wait time and total delay of their travel time. Optimal pedestrian traffic-light equipment, ramp metering, congestion pricing schemes, parking price schemes, Charge for workplace parking, telecommuting programs, etc., can be solutions to the problem of high traffic density. However, in contrast to current human-driven traffic condition, before the introduction of automated vehicles, we may need to redesign urban roads. Therefore, causal-based studies assist transportation planners to make better decisions on redesigning urban roads and applying TDM methods for controlling traffic density and consequently guaranteeing safety of pedestrians based on causal or true effect. In addition, manufacturers should improve the incorporation of pedestrian-to-vehicle communication technologies in automated vehicles so as to improve the process of finding safe gap for crossing.

In the future study, we will compare the structure of machine-learning causal models with the Structural Causal Model (SCM), which in contrast to copula approach, inherently captures causality. In addition, as our analysis is based unconfoundedness assumption, comparing the performance of DML with other causal models in which unobserved variables are considered can be one future direction. In terms of our case study, our data is classified as panel data; however, we did not account for the panel effect and this issue can be investigated in future work. Furthermore, further researchers can make a better comparison between the performance of different machine-learning causal models. Moreover, in terms of pedestrian reaction in a collaborative environment concept, analysis of pedestrian group-crossing behavior and the role of conformity can be a future research direction.

**AUTHOR CONTRIBUTION STATEMENT**
The authors confirm contribution to the paper as follows:
Study conception and design: KK, BF
Data preparation: KK
Methodology and investigation: KK, BF
Coding and estimation: KK
Analysis and interpretation of results: KK, BF
Draft manuscript preparation: KK, BF
Funding and supervision: BF
All authors reviewed the results and approved the final version of the manuscript.

**ACKNOWLEDGEMENTS**
This study is supported by the funds awarded to Dr. Farooq from the Canada Research Chair program and NSERC-Discovery program. The data used in this study were originally collected by Dr. Arash Kalatian under Dr. Farooq's supervision.

Kamal and Farooq    20